\documentclass{article}

\usepackage{amsmath, amsthm} 
\usepackage{amsfonts}
\usepackage{amssymb}
\usepackage{amsthm}

\usepackage[english]{babel}
\usepackage{natbib}
\usepackage{bookmark}
\usepackage{booktabs}

\usepackage{graphicx}

\usepackage{hyperref}

\usepackage{parskip}
\usepackage[noend]{algpseudocode}
\usepackage{algorithm,algorithmicx}

\newcommand{\reals}{\ensuremath{\mathbb{R}}}

\newcommand{\E}{\ensuremath{\mathbb{E}}}
\renewcommand{\P}{\ensuremath{\mathbb{P}}}

\renewcommand{\epsilon}{\varepsilon}
\renewcommand{\v}[1]{\ensuremath{\boldsymbol{\mathrm{#1}}}}
\renewcommand{\b}[1]{\ensuremath{\overline{#1}}}

\newcommand{\cF}{\ensuremath{\mathcal{F}}}

\newcommand{\ind}{{\mathbb{I}}}

\newtheorem{theorem}{Theorem}

\newtheorem{lemma}[theorem]{Lemma}
\newtheorem{definition}{Definition}

\renewcommand{\[}{\begin{equation}}
\renewcommand{\]}{\end{equation}}

\title{Optimal Testing in the Experiment-rich Regime}
\author{
    Sven Schmit, Virag Shah, Ramesh Johari\\
Stanford University}
\date{\today}

\begin{document}

    \maketitle

    \begin{abstract}
    Motivated by the widespread adoption of large-scale A/B testing in
    industry, we propose a new experimentation framework for the setting where
    potential experiments are abundant (i.e., many hypotheses are available to
    test), and observations are costly; we refer to this as the {\em
    experiment-rich regime}.  Such scenarios require the experimenter to
    internalize the opportunity cost of assigning a sample to a particular
    experiment.  We fully characterize the optimal policy and give an algorithm
    to compute it. Furthermore, we develop a simple heuristic that also
    provides intuition for the optimal policy.  We use simulations based on
    real data to compare both the optimal algorithm and the heuristic to other
    natural alternative experimental design frameworks.  In particular, we
    discuss {\em the paradox of power}: high-powered ``classical'' tests can
    lead to highly inefficient sampling in the experiment-rich regime.
\end{abstract}

    \section{Introduction}
\label{sec:intro}

In modern A/B testing (e.g., for web applications), it is not uncommon to find
organizations that run hundreds or even thousands of experiments at a time
\citep{kaufman2017democratizing, google_experiments, Kohavi2009OnlineEA,
bakshy2014designing}.  Increased computational power and the ubiquity of
software have made it easier to generate hypotheses and deploy experiments.
Organizations typically continuously experiment using A/B testing.  In
particular, the space of potential experiments of interest (i.e., hypotheses
being tested) is vast; e.g., testing the size, shape, font, etc., of page
elements, testing different feature designs and user flows, testing different
messages, etc.  Artificial intelligence techniques are being deployed to help
automate the design of such tests, further increasing the pace at which new
experiments are designed (e.g., Sensei, Adobe's A/B testing
product, is being used in Adobe Target).\footnote{\url{https://www.adobe.com/marketing-cloud/target.html}}

This abundance of potential experiments has led to an interesting phenomenon:
despite the large numbers of visitors arriving per day at most online web
applications, organizations need to constantly consider the most efficient way
to allocate these visitors to experiments.  For many experiments, baseline
rates may be small (e.g., a low conversion rate), or more generally effect
sizes may be quite small even relative to large sample sizes.  For example,
large organizations may be seeking relative changes in a conversion rate of
$0.5$\% or less, potentially necessitating millions of users allocated to a
single experiment to discover a true effect.  (See \cite{google_experiments,
google_experiments_preso, deng2013improving} and \cite{azevedo2018abtesting},
where these issues are discussed extensively.) Since organizations have a
plethora of hypotheses of interest to test, there is a significant {\em
opportunity cost}: they must constantly trade off allocation of a visitor to a
current experiment against the potential allocation of this visitor to a new
experiment.

In this paper, we study a benchmark model with the feature that experiments are
abundant relative to the arrival rate of data; we refer to this as the {\em
experiment-rich regime}.  A key feature of our analysis is the impact of the
opportunity cost described above: whereas much of optimal experiment design
takes place in the setting of a {\em single} experiment, the experiment-rich
regime fundamentally requires us to trade off the potential for discoveries
across {\em multiple experiments}.  Our main contribution is a complete
description of an optimal discovery algorithm for our setting; the development
of an effective heuristic; and an extensive data-driven simulation analysis of
its performance against more classical techniques commonly applied in
industrial A/B testing.

We present our model in Section~\ref{sec:model}.  The formal setting we
consider mimics the setting of most industrial A/B testing contexts. The
experimenter receives a stream of observational units and can assign them to an
infinite number of possible experiments, or {\em alternatives}, of varying
quality ({\em effect size}).  We consider a Bayesian setting where there is a
prior over the effect size of each alternative, which is natural in a setting
with an infinite number of experiments.

We focus on the objective of finding an alternative that is at least
as good as a given threshold $s$ as fast as possible.  In particular,
we call an alternative a {\em discovery} if the posterior probability
that the effect is greater than $s$ is at least $1 - \alpha$, and the
goal is to minimize the expected time per discovery.  This is a
natural criterion: good performance requires finding an alternative
that is actually delivering practically significant effects (as
measured by $s$).  Adjusting $s$ and $\alpha$ allows the experimenter
to trade off the quality and quantity of discoveries made.  Note that
under this criterion any optimal policy is naturally incentivized to
find the ``best'' experiments, because the discovery criterion is
easiest to be met for those alternatives.

In Section~\ref{sec:optimal} we present an optimal policy for allocation of
observations to experiments.  Since observations arrive sequentially, the
problem can be equivalently formulated as minimizing the cumulative number of
observations until until a discovery is made.  We characterize a dynamic
programming approximation of this problem, and show this method converges to
the optimal policy in an appropriate sense.  We also develop a simple heuristic
that approximates and provides insight into the optimal policy.

In Section~\ref{sec:baseball} we use data on baseball players' batting averages
as input data for a simulation analysis of our approach.  Our simulations
demonstrate that our approach delivers fast discovery while controlling the
rate of false discoveries; and that our heuristic approximates the optimal
policy well.  We also use the simulation setup to compare our method to
``classical'' techniques for discovery in experiments (e.g., hypothesis
testing).  This comparison reveals the ways in which classical methods can be
inefficient in the experiment-rich regime.  In particular, there is a {\em
paradox of power}: efficient discovery can often lead to low power in a
classical sense, and conversely high-powered classical tests can be highly
inefficient in maximizing the discovery rate.

Due to space constraints, all proofs are given in the appendix.

\subsection{Related work}

The literature on sequential testing goes back many decades.
Originally, \citet{wald1948optimum} propose an optimal test, the
sequential probability ratio test, or SPRT for short, for testing a simple hypothesis.
\citet{chernoff1959sequential} studies the asymptotics of experimentation
with two hypothesis tests and how to assign observations.
\cite{lai1988nearly} proposes a class of Bayesian sequential tests with a composite
alternative for an exponential family of distributions.
For a more thorough overview of sequential testing, we refer the interested
reader to \cite{siegmund2013sequential}, \cite{wetherill1986sequential},
\cite{shiryayev1978optimal} and \cite{lai1997optimal}.
None of these approaches consider the opportunity cost associated with having
multiple experiments.

Recently, there has been an increased interest in sequential testing due to the
rise in popularity of A/B testing \citep{Deng2017TrustworthyAO,
kaufman2017democratizing, Kharitonov2015SequentialTF}, and the ubiquity of peeking
\citep{Johari2017PeekingAA, Balsubramani2016SequentialNT, Deng2016ContinuousMO}.
A recent paper by \cite{azevedo2018abtesting} discusses how the tails of the effect
distribution affect the assignment strategy of observations to experiments and
complements this work.

There is also a strong connection to the multi-armed bandit literature
\citep{gittins2011multi,Bubeck2012RegretAO}, especially the pure exploration
problem \citep{Bubeck2009PureEI, Jamieson2014lilU, russo2016simple}, where the
goal is to find the best arm.  The case with infinitely many arms is
studied by \cite{Carpentier2015SimpleRF, chaudhuri2017pac, Aziz2018PureEI}.
\cite{Locatelli2016AnOA} studies the setting of finding the set of arms (out of
finitely many) above a given threshold in a fixed time horizon.

Methods to control of the false discovery rate in the sequential hypothesis
setting are discussed by \cite{Foster2007AlphainvestingAP},
\cite{Javanmard2016OnlineRF} and \cite{ramdas2017online}.  The connection
between with multi-armed bandits is made by \cite{yang2017framework}.  However,
the Bayesian framework we propose does not require multiple testing
corrections.

The heavy-coin problem \citep{Chandrasekaran2012FindingTM,
Malloy2012QuickestSF, Jamieson2016ThePO, Lai2011QuickestSO} is another closely related research area. Here,
a fraction of coins in a bag is considered heavy, while most are light.
The goal is to find a heavy coin as quickly as possible.
These approaches rely on likelihood ratios, as there are only two alternatives,
and there is a connection to the CUSUM procedure \citep{page1954continuous}.
The approaches mentioned above all consider the same problem as we do in this work,
albeit for testing two alternatives against each other.

Optimal stopping rules have been studied extensively, often under the umbrella
of the secretary problem \citep{freeman1983secretary, samuels1991secretary}.
There, the focus is on comparing across alternatives.

    \section{Model and objective}
\label{sec:model}

In this section we describe the model we study and the objective of the experimenter.

{\bf Experiments}.  We consider a model with an infinite number of experiments, or alternatives, indexed
by $i\in\{1,2,\ldots\}$.  Each experiment is associated with a parameter $\mu_i \in M \subset
\reals$ drawn independently from a common (known) prior $\pi$ that completely characterizes
the distribution of outcomes corresponding to that experiment.  Throughout our analysis, the experimenter is interested in experiments with higher values of $\mu_i$.

{\bf Actions and outcomes}.  At times $t=1,2,\ldots$, the experimenter selects an alternative $I_t$ and observes an
independent outcome $X_t$ drawn from a distribution $F(\mu_{I_t})$.
Note, in particular, that opportunities for observations arrive in a sequential, streaming fashion.  We also assume that observations are independent across experiments.

We assume that $F(\mu_i)$ is described by a single parameter natural exponential
family, i.e. the density for an observation can be written as:
\[
    f_X(x \mid \mu) = h(x) \exp \left( \mu S(x) - A(\mu) \right),
\]
for known functions $S$, $h$, and $A$.
Let $S_t^i = \sum_{t: I_k = i} S(X_t)$ be the canonical sufficient statistic for experiment $i$ at time $t$.  Note that in particular, this model includes the conjugate normal model with known variance and
the beta-binomial model for binary outcomes.

{\bf Policies}.  Let $\cF_t = \sigma\{X_1, I_1, X_2, I_2, \ldots, X_t, I_t\}$ denote the $\sigma$-field
generated.  A {\em policy} is a mapping from $\cF_t$ to experiments.

{\bf Discoveries}.  The experimenter is interested in finding \emph{discoveries}, defined as follows.
\begin{definition}[Discovery]
We say that alternative $i$ is a \emph{discovery} at time $t$, given $s$ and $\alpha$, if
\begin{equation}
\label{eq:discovery}
        \P(\mu_i < s \mid \cF_t) < \alpha.
\end{equation}
\end{definition}
Here $s$ and $\alpha$ are parameters that capture the experimenter's
preferences, i.e., the level of aggressiveness and risk that she is willing to
tolerate.  (Note that this is more
stringent than the related false discovery rate guarantees \citep{Benjamini2007ControllingTF}.)

We assume that the prior satisfies $\P(\mu_i < s \mid \emptyset) \in (\alpha, 1)$ to avoid
the trivial scenarios that all or none of the alternatives is a discovery before trials begin.

{\bf Objective: Minimize time to discovery.}  As motivated in the introduction, informally the objective is to find discoveries as fast as possible.  We formalize this as follows: The goal of the experimenter is
to design a policy (i.e., an algorithm to match observations to experiments)
such that the number of observations until the first discovery is minimized.

In particular, define the time to first discovery $\tau$ as:
\[
    \tau = \min \{ t : \text{there exists } i^* \text{ such that } \P(\mu_{i^*} < s \mid \cF_t ) < \alpha \}.
\]

Then the goal is to {\em minimize $\E[\tau]$} over all policies. Given this
goal, the only decision the experimenter needs to make at each point in time
till the first success is whether to reject the current experiment or to
continue with it.


{\bf Discussion}.  We conclude with two remarks regarding our model.

(1) {\em Posterior validity}.  Note that at the (random) stopping time $\tau$, the posterior is computed based on the potentially adaptive matching policy used by the experimenter. The following lemma shows that when the experimenter computes the posterior and decides to stop the experiment at time $t$ when the condition $\P(\mu_{i^*} < s \mid \cF_t )$ is met, the decision to stop does not invalidate the discovery.

\begin{lemma}
\label{thm:validity}
    The posterior for the discovered experiment $i^*$ at time $\tau$ satisfies
    \[
        \P(\mu_{i^*} < s \mid \cF_\tau) < \alpha
    \]
    almost surely.
\end{lemma}

(2) {\em Fixed cost per experiment}.  In some scenarios, starting a new experiment has a cost; e.g., there may be a cost to implementing a new variant,
or results may need to be analyzed on a per experiment basis.
We can incorporate such a cost in the objective, and our results and approach generalize accordingly.  Formally, let $c$ be the cost of starting a new experiment, and let
$m_t = | \{ i : \exists t' \le t: I_{t'} = i \} |$ be the cumulative number of
matched experiments up to time $t$.  We can include the per experiment cost by considering instead the problem of {\em minimizing $\E[\tau + c m_\tau]$}.

    \section{Optimal policy}
\label{sec:optimal}

In this section, we characterize the structure of the optimal policy,
show that it can be approximated arbitrarily well by considering a truncated
problem, and give an algorithm to compute the optimal policy of the truncated problem.
Finally, we present a simple heuristic that approximates the optimal policy remarkably
well.

\subsection{Sequential policies}

We start with a key structural result that simplifies the search for an optimal policy.  The following lemma shows that we can focus on policies that only consider experiments {\em sequentially}, in the sense that once a new experiment is being allocated observations, no previous experiment will ever again receive observations.

\begin{lemma}
    \label{thm:sequential}
    There exists an optimal policy such that $I_{t+1} \ge I_t$ for all $t$ almost surely.
\end{lemma}

This result hinges on three aspects of our model: experiments are independent
of each other, with identically distributed effects $\mu_i$; there are an infinite number of experiments available; and observations arrive in an infinite stream.  As a consequence, all experiments are {\em a priori} equally viable, and {\em a posteriori} once the experimenter has determined to stop allocating observations to an experiment, she need never consider it again.

Note in particular that this lemma also reveals that any optimal policy for the first discovery also straightforwardly minimizes the expected time until the $k$'th discovery, for any $k$.

\subsection{Reformulating the optimization problem}
\label{sec:simplified}

Based on Lemma~\ref{thm:sequential}, we can reformulate and simplify the
optimization problem faced by the experimenter as a sequential decision
problem, where the only choice is whether or not to continue testing the {\em
current} experiment.

We abuse notation to describe this new perspective.  Let $\mu$ denote the
effect size of the current experiment.  In particular, let $X_n$ be the $n$'th
observation; let $\cF_n$ be the $\sigma$-field generated by observations
of the current experiment $(X_1, \ldots, X_n)$.
Let $S_n = \sum_{k = 1}^n S(X_k)$ denote the canonical sufficient
statistic at state $n$.
The {\em state} of the sequential decision problem is $(n, S_n)$, the
number of observations and the sufficient statistic of the current experiment.

If $(n, S_n)$ has the property that $\P(\mu < s | S_n) < \alpha$, then a
discovery has been found and so the process stops.  The following lemma
shows that this discovery criterion induces an {\em acceptance region} on the
sufficient statistic $S_n$, i.e., a sequence of thresholds $a_n$ such the
current experiment is a discovery when $S_n \ge a_n$.

\begin{lemma}
  \label{thm:acceptance}
  There exists a sequence $\{a_n\}_{n=1}^\infty$ such that
  $\P(\mu < s \mid S_n) < \alpha$ if and only if $S_n > a_n$.
\end{lemma}

If $S_n < a_n$, then the experimenter can make one of two decisions:

\begin{enumerate}
  \itemsep0em
  \item {\em Continue} (i.e., collect one additional observation on the current experiment); or
  \item {\em Reject} (i.e., quit the current experiment and collect the first observation of a new experiment).
\end{enumerate}
If {\em Continue} is chosen, the state updates to $(n+1, S_{n+1})$.  If {\em Reject} is
chosen, the state changes to $(1, S_1)$ (where $S_1$ is an independent draw of the sufficient statistic after the first observation); and in either case, the process continues.

The goal of the experimenter is to minimize the expected time until the
observation process stops, i.e., until a discovery is found.  Let $V(n, S_n)$
be this minimum, starting from state $(n, S_n)$.  Then the Bellman equation for
this process is as follows:
\begin{align}
V(n, S_n) &= 0, \ \ S_n \ge a_n; \label{eq:bellman1} \\
V(n, S_n) &= 1 + \min \left \{\E[V(n+1, S_{n+1}) | S_n], \E[V(1, S_1)] \right \}, \ \ S_n \leq a_n \ \ n \ge 1. \label{eq:bellman2}
\end{align}
The first line corresponds to the case where $S_n$ is in the acceptance region,
i.e., the process stops.  In the second line, we consider two possibilities:
continuing incurs a unit cost for the current observation, plus the expected
cost from the state $(n+1, S_{n+1})$; rejecting resets the state with no cost
incurred.  The optimal choice is found by minimizing between these
alternatives.
The expected number of samples $T^*$ till a discovery satisfies $T^* = 1 + \E[V(1, S_1)]$.

\subsection{Characterizing the optimal policy}

The following theorem shows that an optimal policy for the dynamic programming
problem \eqref{eq:bellman1}-\eqref{eq:bellman2} can be expressed using a
sequence of rejection thresholds on the sufficient statistic.  That is, for
each $n$ there is an $r_n$ such that it is optimal to {\em Continue} if $S_n
\ge r_n$, and to {\em Reject} if $S_n < r_n$.

\begin{theorem}
  \label{thm:optimal}
There exists an optimal policy for \eqref{eq:bellman1}-\eqref{eq:bellman2}
described by a sequence of rejection thresholds $\{r_n\}_{n=1}^\infty$
  such that, after $n$ observations, {\em Reject} is declared if $S_n < r_n$, {\em Continue} is declared if $r_n \leq S_n \leq a_n$, and the process stops with a discovery if $S_n > a_n$.
\end{theorem}

The remainder of the section is devoted to computing the optimal sequence of
rejection thresholds.

\subsection{Approximating the optimal policy via truncation}

In order to compute an optimal policy, we consider a {\em truncated} problem.
This problem is identical in every respect to the problem in Section
\ref{sec:simplified}, except that we consider only policies that must choose
{\em Reject} after $k$ observations.  We refer to this as the {\em
$k$-truncated} problem.

Let $V_k(n, S_n)$ denote the minimum expected cumulative time to discovery for
the $k$-truncated problem, starting from state $(n, S_n)$. The Bellman equation
is nearly identical to \eqref{eq:bellman1}-\eqref{eq:bellman2}, except that now
$V_k(k, S_k) = 1+\E[V_k(1, S_1)], S_k \leq a_k$, and we add
the additional constraint that $n<k$ to \eqref{eq:bellman2}.
We have the following proposition.

\begin{theorem}
  \label{thm:truncated}
  There exists an optimal policy  for the $k$-truncated problem described by a sequence of rejection thresholds
  $\{ r_n^k \}_{n=1}^\infty$ such that, after $n$ observations, {\em Reject} is declared if $S_n < r_n^k$, {\em Continue} is declared if $r_n^k \leq S_n \leq a_n$, and {\em Accept} is declared is if $S_n > a_n$.

Further, let $T_k^* = \E[V_k(1, S_1)] + 1$ be the optimal expected number of
observations until a discovery is made.  Then for each $n$, $r_n^k \to r_n$ as
$k \to \infty$; and $T_k^* \to T^*$ as $k \to \infty$.
\end{theorem}

\subsection{Computing the truncated optimal policy}
\label{sec:computing}

The truncated horizon brings us closer to computing an optimal policy,
but it is still an infinite horizon dynamic programming problem.  
In this section we show instead that we can compute the truncated optimal policy by iteratively solving a single-experiment truncated problem with a fixed
rejection cost $\kappa$.
Let $W_k(n, S_n|\kappa)$ be the optimal expected cost for this problem starting from state $(n, S_n)$.  We have the following Bellman equation.
\begin{align}
W_k(n, S_n|\kappa) &= 0,\ \ S_n > a_n;\label{eq:bellman_single1} \\
W_k(k, S_k|\kappa) &= \kappa,\ \ S_k \leq a_k;\label{eq:bellman_single2} \\
W_k(n, S_n|\kappa) &= 1+\min \left \{\E[W_k(n+1, S_{n+1} | \kappa) | S_n], \kappa \right \}, \ \ n < k, S_n \leq a_n. \label{eq:bellman_single3}
\end{align}

For any terminal cost $\kappa$, this dynamic programming problem is easily solved using
backward induction to find the rejection boundaries.  The following theorem shows how we can use this solution to find an optimal policy to the truncated problem.

\begin{theorem}
  \label{thm:onestep}
  If $\kappa = T_k^*$, then the optimal policy for \eqref{eq:bellman_single1}-\eqref{eq:bellman_single3} with rejection thresholds $\b r_n^k$ found by backward induction
  satisfies $\b r_n^k = r_n^k$ for all $n \le k$.
  Furthermore, let $f(\kappa) = 1+\E[W_k(1, S_1 | \kappa)]$ be the optimal cost. Then if $\kappa > T_k^*$, $f(\kappa) < \kappa$, and
  if $\kappa < T_k^*$, then $f(\kappa) > \kappa$.
\end{theorem}

Thus, to find approximately optimal rejection thresholds, select $k$ suitably large,
and start with an arbitrary $\kappa$. Then iteratively compute the corresponding
thresholds $\b r_n^k$ and the cost $f(\kappa)$, using bisection to converge on $T_k^*$, and thus
the corresponding optimal thresholds.

We note that the same program we have outlined in this section
can be used to compute an optimal policy with a per experiment fixed cost $c$,
by using rejection cost $\kappa + c$ instead of $\kappa$.
Empirically, this leads to only slightly lower rejection
thresholds; due to space constraints, we omit the details.



    \subsection{Heuristic approximation}
\label{sec:heuristic}

We have seen that the optimal policy is easy to approximate by solving dynamic
programs iteratively. However, this does not give us direct insight into the structure
of the solution, and in certain cases a quick rule-of-thumb that provides
an approximate policy might be all that is required. In this section, we show
that there exists a simple heuristic that performs remarkably well.

\begin{figure}
\centering
\includegraphics[width=0.7\textwidth]{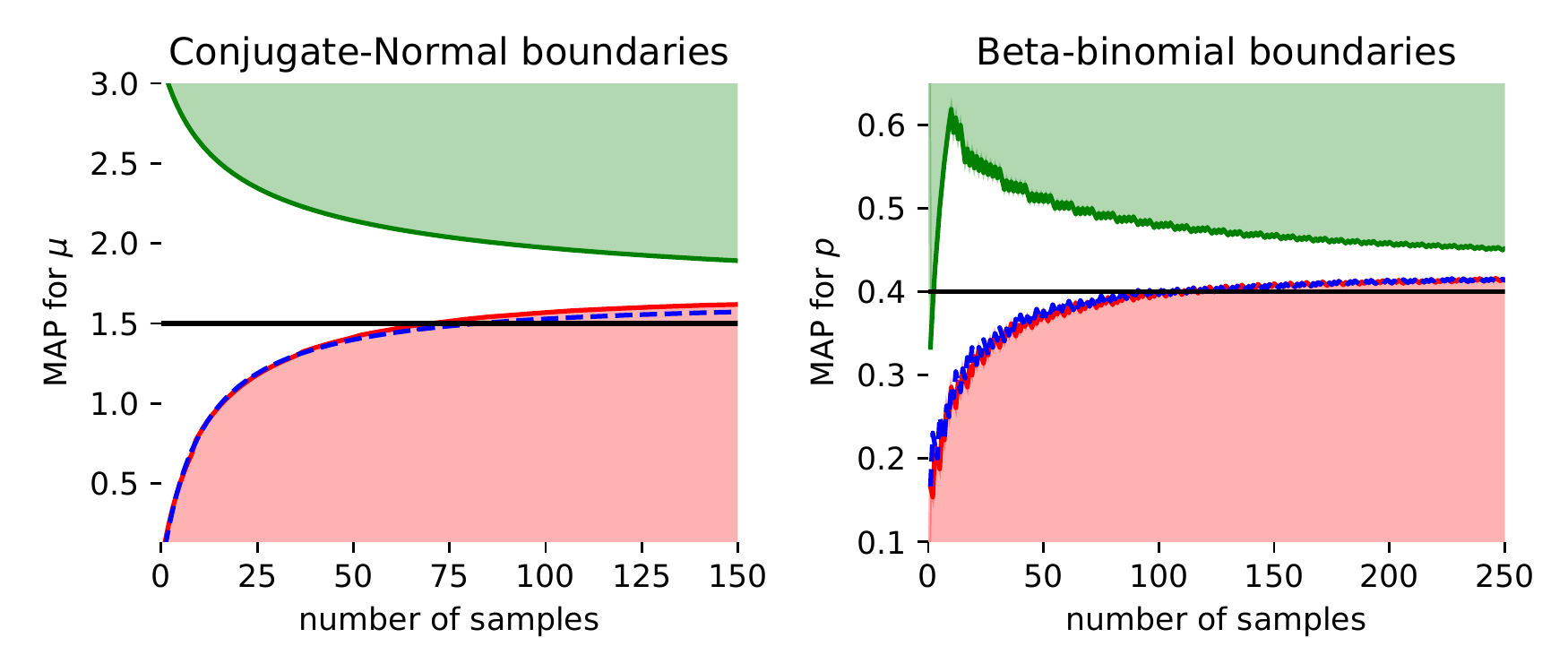}
\caption{Acceptance and rejection regions for the conjugate Normal and the Beta-Binomial
models. The dashed blue line gives the heuristic rejection boundary, while the red line
corresponds to the optimal rejection thresholds. Note that the
boundaries are shown in terms of the MAP estimate.}
\label{fig:boundary}
\end{figure}

The approximate rejection boundary at time $n$ is found as follows. Let
$\hat \mu$ be the MAP estimate of $\mu$ for sufficient statistic $S_{n+T^*} = a_{n+T^*}$.
Then reject the current experiment if
$S_n$ is not plausible under $\hat \mu$. That is, the heuristic boundary $\b r_n$
is, for a suitably chosen $\beta$,
\[
    \P(S_n \le \b r_n \mid \mu = \hat \mu) = \beta.
\]
Of course, this heuristic is not practical as is, as in general we do not know $T^*$ unless
we compute the optimal policy. But often $a_{n+t}$ varies only little in $t$ so a
reasonable approximate choice $T_h$ is sufficient.
In Figure~\ref{fig:boundary} we plot the discovery and rejection boundaries,
along with the heuristic outlined above (with $T_h = T^*$), for the normal and Bernoulli models.

The heuristic and optimal policies clearly exhibit aggressive rejection regions, cf.~Figure \ref{fig:boundary}.
THe interpretation is as follows: to continue sampling from the current experiment, we do not just want its quality
to be $s$, but substantially better than $s$, since $a_n > s$ for all $n$.
If not, it would take too many additional observations to verify the discovery.

    \section{Case study: baseball}
\label{sec:baseball}

We now empirically analyze our testing framework based on a simulation with
baseball data.
First, we demonstrate empirically that the proposed algorithm leads
to fast discoveries,
and behaves differently from traditional testing approaches.
Second, we show that the rule-of-thumb heuristic performance is close to that
of the optimal policy.\footnote{Code to replicate results can be found at \url{https://github.com/schmit/optimal-testing-experiment-rich-regime}.}

\textbf{Data}
We use the baseball dataset with pitching and hitting statistics from 1871
through 2016 from the Lahman R package.
The number of At Bats (AB) and Hits (H) is collected for each player,
and we are interested in finding players with a high batting average, defined
as $b_i = \text{Hits}_i / \text{At Bats}_i$.
We consider players with at least $200$ At Bats, which leaves a total
of 5721 players, with a mean of about 2300 At Bats.
In the top left of Figure~\ref{fig:baseball_results}, we plot the histogram of batting
averages, along with an approximation by a beta distribution (fit via method of moments).
We note that these fit the data reasonably well, but not perfectly.
This discrepancy helps us evaluate the robustness to a misspecified prior.


\textbf{Simulation setup}
To construct the testing problem, we view the batters as alternatives, with empirical batting average $b_i$ of batter $i$ treated as ground truth.  We want to find alternatives with $b_i > s$.
We draw a Bernoulli sample of mean $b_i$ to simulate an observation from
alternative $i$.\footnote{Based on sequential batting data from the 2014-2018
seasons there is no evidence for strong correlation between at-bats.} These
samples are then used to test whether $b_i > s$.  We set $\alpha = 0.05$, and
vary $s$ between $0.25$ and $0.32$.  For each simulation, we iterate through
each batter and repeat it 1000 times to reduce variance.
This allows us to compare methods fairly, ensuring that
each procedure is run on exactly the same test cases.

\subsection{Benchmarks}

To assess performance, we compare several testing procedures.
Note that the non-traditional setup of our testing framework does not allow
for easy comparison with other methods, in particular frequentist approaches,
as they give different guarantees.
Thus, we restrict attention to Bayesian methods that provide the same error guarantee.
All of the benchmarks use the same beta prior computed above.

\paragraph{Optimal policy}
First we study the optimal policy based on the beta-binomial model,
computed using the bisection and
backward induction approach in Section~\ref{sec:computing},
where we truncate after $k = 5000$ samples.

\paragraph{Heuristic policy}
Next, we include the heuristic rejection thresholds that approximate the optimal policy for truncation $k = 5000$ samples.
The heuristic policy requires setting two parameters:
$T_h$, i.e., how far to look into the future to find the acceptance boundary, which
is ideally set close to $T^*$; and the rejection region $\beta$.
To demonstrate the the insensitivity to $T^*$, we use $T_h=2000$ and $\beta=0.2$ for all simulations.
(Note that $T^*$ varies dramatically as we change the threshold
$s$.

\paragraph{Fixed sample size test}
Our next benchmark is a simple fixed sample size test.
For each experiment, we gather $N$ observations, and claim a discovery
if $P(\mu_i < s \mid Y_i) < \alpha$ where $Y_i$ is the number of Hits of alternative (batter) $i$.
We focus our attention on using $N=1000$ samples per test, as this seems
to perform best when compared to other sample sizes, but any differences are immaterial
for our conclusions.

\paragraph{Fixed sample size test with early stopping}
This benchmark is similar to the fixed sample size test, except that
we stop the experiment early if the discovery criterion is met.
Thus, we can quantify the gains from being able to discover early.


\paragraph{Bayesian sequential test}
Now we consider a sequential test that also rejects early.
In particular, we reject the current experiment if
$
    \P(b_i > s \mid \v S_t^i) < \beta.
$
We also reject an alternative after $4000$ samples.
This approach also requires careful tuning of $\beta$.
In particular, if $\beta$ is too large, say larger than the prior probability
$\P_0(b_i > s)$, then the test is too aggressive and rejects all alternatives outright.
Instead, we found empirically that setting $\beta = 0.9 \P_0(b_i > s)$ leads to good performance
across all values of $s$.

\subsection{Results}

\begin{figure}
    \centering
    \includegraphics[width=0.85\textwidth]{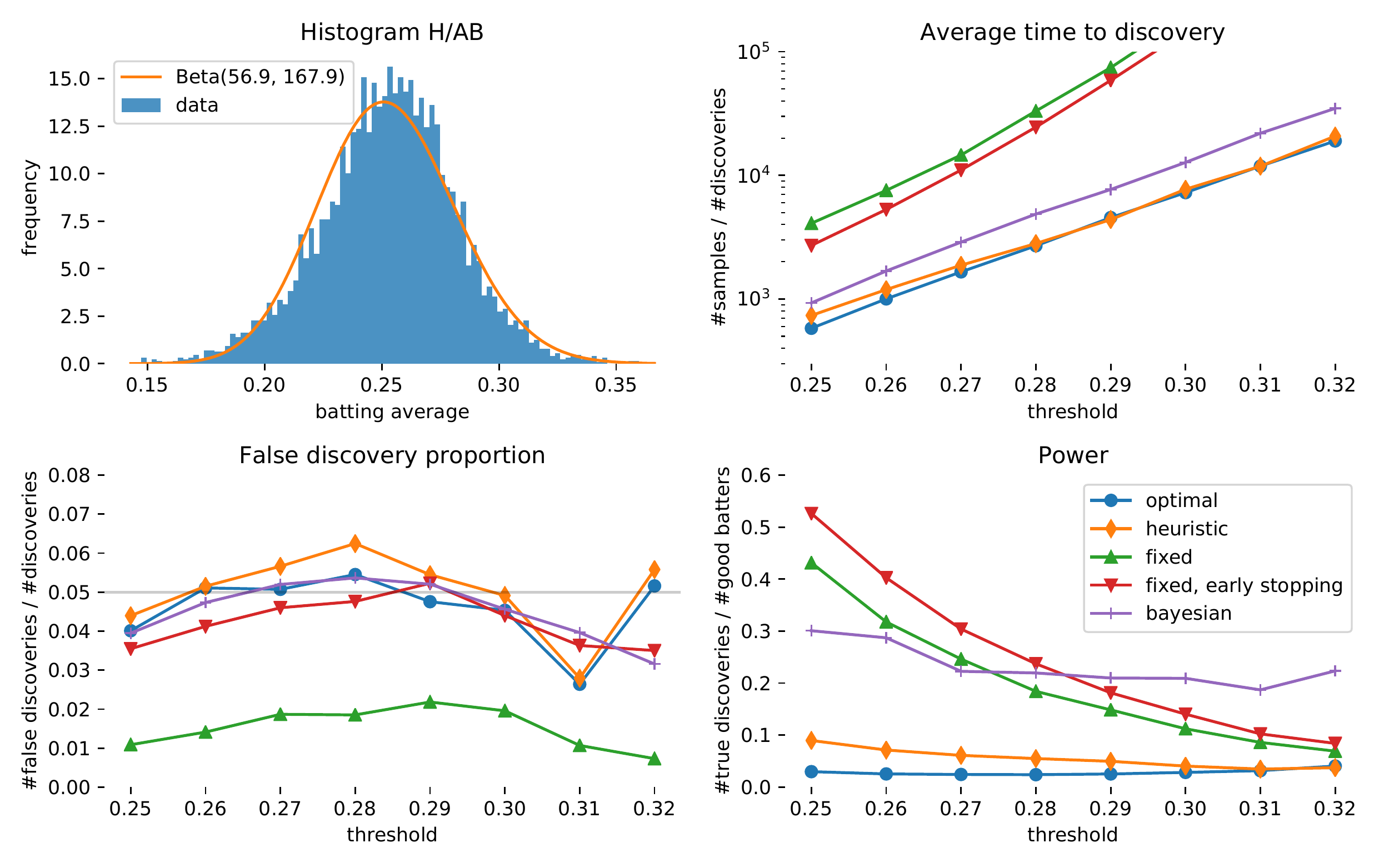}
    \caption{Top left: histogram of batting averages. Top right: Efficacy of of
    algorithms. Bottom left: Plot the false discovery proportion across thresholds.
    Bottom right: Plot of the empirical power of algorithms.  Note the {\em paradox of power} effect:
    the most efficient algorithms have low power.}
    \label{fig:baseball_results}
\end{figure}

\paragraph{Average time to discovery}

The average number of observations until a discovery is shown in the top right
plot of Figure~\ref{fig:baseball_results}.
As expected, the fixed sample test performs worst.
Early stopping leads to slightly better performance, but this method is still
not effective as most of the gains come from early rejection.  The Bayesian
sequential test demonstrates this effect and shows substantial gains over the
fixed tests. The heuristic policy, despite lack of parameter
tuning, performs very well, essentially matching the performance of the optimal
algorithm for most thresholds.

\paragraph{False discovery proportion and robustness}
Next, we compare the {\em false discovery proportion} (FDP) \citep{Benjamini2007ControllingTF}, i.e., the fraction of discoveries that in fact had true $b_i < s$.
If the prior is correctly specified, the methods we consider satisfy
$\E(\textup{FDP}) \le \alpha$.
Indeed, we observe that the guarantee holds for most thresholds and algorithms in
the bottom left plot of Figure~\ref{fig:baseball_results}.
There is some minor exceedance of the FDP for thresholds around $s\approx 0.28$,
which can be explained by the fact that the prior does not fit the empirical batting averages
perfectly.
Since there are few rejections for thresholds beyond $s=0.3$, the FDP estimate has higher variance in that regime.
Across all simulations, the optimal policy has an FDP of $0.048 < \alpha$.
Finally, we see that the lack of early stopping makes the fixed test rather conservative.

\paragraph{The paradox of power}

Finally, we compare {\em power}, i.e., the fraction of alternatives $i$ with $b_i > s$ that are declared a discovery.  Power comparisons across the algorithms are  plotted in the bottom right of Figure~\ref{fig:baseball_results}.
The most surprising insight from the simulations is the \emph{paradox of power}.
Algorithms that are effective have very low power.
This is counter-intuitive: how can algorithms that make many discoveries have only
a small chance of picking up true effects?
The main driver of good performance for an algorithm is the ability to quickly reject
unpromising alternatives.
Some unpromising alternatives are ``barely winners'': i.e., $b_i$ is only slightly above $s$.  In the experiment-rich regime, such alternatives should be rejected quickly, because it takes too many observations
to get enough concentration around the posterior to claim a discovery.
This effect leads to low power, but fast discoveries.

\begin{figure}
    \centering
    \includegraphics[width=\textwidth]{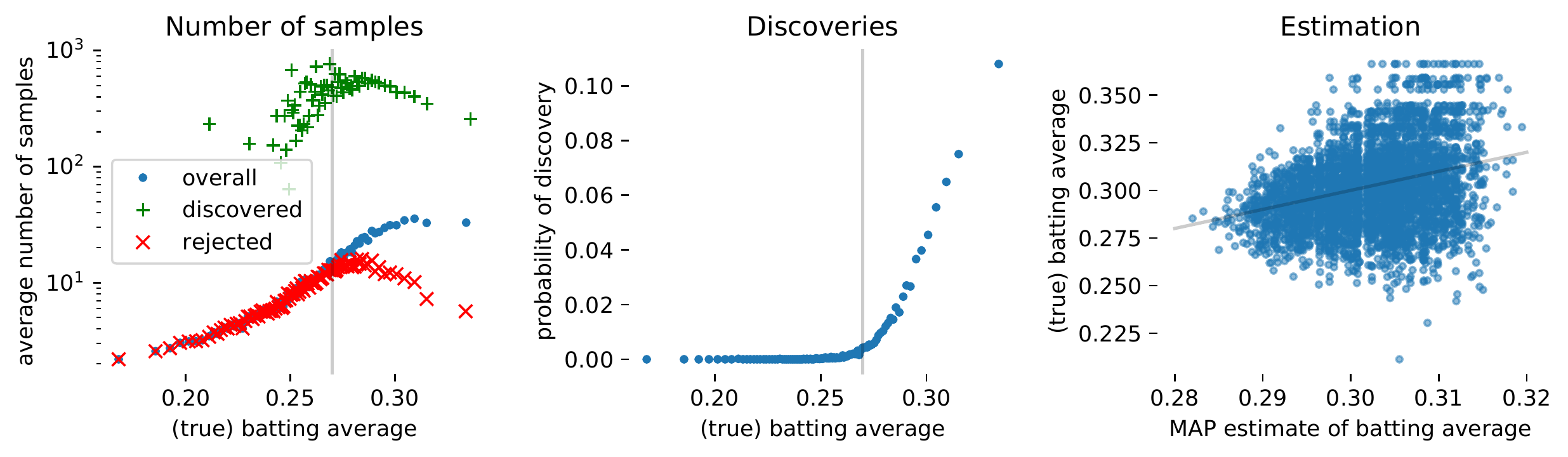}
    \caption{Analysis of optimal policy. The left plot shows average number of samples
    for rejections and discoveries. The center plot shows the fraction of discoveries.
    The right plot shows MAP estimates of effect sizes.}
    \label{fig:results_optimal}
\end{figure}

\paragraph{Characteristics of the optimal policy}
We consider the outcomes of individual tests for the optimal
algorithm ($s=0.27$) in Figure~\ref{fig:results_optimal}.
The average number of samples for rejected
alternatives is very small while it is much larger for
discovered alternatives.  We also note the concave shape for the discovered
alternatives, that seems to peek around $0.285$.  When the batting average is
larger, the algorithm is able to detect the effect with fewer samples, and when
the batting average is lower, there are a few lucky streaks that lead to a
(false) discovery.

The probability of discovery is low across all
batting averages, but increases sharply beyond a batting average of $0.28$,
rather than around $s=0.27$.  As noted before,  the optimal policy tries
to avoid effects that are close to the threshold.

Finally, the MAP estimates for the batting averages of discovered batters.  It
illustrates a known but important fact that the parameter of discovered
alternatives are quite poor.  If estimation of effects is important, the
experimenter ought to obtain more samples for the discovered alternatives.

    \section{Conclusion}

We consider an experimentation setting where observations are costly, and there
is an abundance of possible experiments to run ---  an increasingly prevalent
scenario as the world is becoming more data-driven.  Based on backward
induction, we can compute an approximately optimal algorithm that allocates
observations to experiments such that the time to a discovery is minimized.
Simulations validate the efficacy of our approach, and also reveal discuss the
\emph{paradox of power}: there is a tension between high-powered tests, and
being efficient with observations.

Our paradigm has several additional practical benefits.  First, we can leverage
knowledge across experiments through the prior.  Second, adaptive matching of
observations to experiments does not preclude valid inference, and thus
outcomes can thus continuously be monitored.  Finally, the framework also
provides an easy ``user interface'': it directly incorporates the desired
effect size, and leads to guarantees that are easy to explain to non-experts.

\paragraph{Further directions}

The framework assumes there is a common prior among alternatives, and this
allows us to view every rejection as a renewal. If experiments have different priors,
then the order at which experiments are chosen matters. This is also
true when the costs of observations or starting experiments differ across experiments.

Furthermore, we assume the prior is known. The experimenter can take an empirical Bayes
approach similar to our simulations before starting the experiments, but data
gathered while the optimal policy is running distorts estimates of the prior.

We briefly touched upon the independence assumption across observations of a single
experiment, and showed that for baseball data this does not lead to problems, but
in other use cases time variation, e.g. novelty effects, might play a bigger role
and need to be encoded into the framework.
One way to incorporate such effects, along with suitably chose covariates
that can reduce the variance of testing and thereby improving the time till
discoveries are Bayesian generalized linear models.

Finally, we assume that experiments are independent. In certain settings
the results of one experiment can affect future experiments, or there might
be correlations between outcomes of experiments. Incorporating these effects is
non-trivial and beyond the scope of this work.

    \section{Acknowledgements}

The authors would like to thank
Johan Ugander,
David Walsh,
Andrea Locatelli, and
Carlos Riquelme for their suggestions and feedback.
This work was supported by the Stanford TomKat Center, and by the National Science Foundation under
Grant No. CNS-1544548 and CNS-1343253.
Any opinions, findings, and conclusions or recommendations expressed in this
material are those of the author(s) and do not necessarily reflect the views of the National Science
Foundation.

    \bibliographystyle{plainnat}
    \bibliography{bib/bibliography}

\begin{thebibliography}{41}
\providecommand{\natexlab}[1]{#1}
\providecommand{\url}[1]{\texttt{#1}}
\expandafter\ifx\csname urlstyle\endcsname\relax
  \providecommand{\doi}[1]{doi: #1}\else
  \providecommand{\doi}{doi: \begingroup \urlstyle{rm}\Url}\fi

\bibitem[Azevedo et~al.(2018)Azevedo, Deng, Olea, Rao, and
  Weyl]{azevedo2018abtesting}
Eduardo~M Azevedo, Alex Deng, Jose~Montiel Olea, Justin~M Rao, and E~Glen Weyl.
\newblock A/b testing.
\newblock In \emph{Proceedings of the Nineteenth ACM Conference on Economics
  and Computation}. ACM, 2018.

\bibitem[Aziz et~al.(2018)Aziz, Anderton, Kaufmann, and Aslam]{Aziz2018PureEI}
Maryam Aziz, Jesse Anderton, Emilie Kaufmann, and Javed~A. Aslam.
\newblock Pure exploration in infinitely-armed bandit models with
  fixed-confidence.
\newblock In \emph{ALT}, 2018.

\bibitem[Bakshy et~al.(2014)Bakshy, Eckles, and Bernstein]{bakshy2014designing}
Eytan Bakshy, Dean Eckles, and Michael~S Bernstein.
\newblock Designing and deploying online field experiments.
\newblock In \emph{Proceedings of the 23rd international conference on World
  wide web}, pages 283--292. ACM, 2014.

\bibitem[Balsubramani and Ramdas(2016)]{Balsubramani2016SequentialNT}
Akshay Balsubramani and Aaditya Ramdas.
\newblock Sequential nonparametric testing with the law of the iterated
  logarithm.
\newblock \emph{CoRR}, abs/1506.03486, 2016.

\bibitem[Benjamini and Hochberg(2007)]{Benjamini2007ControllingTF}
Author~Yoav Benjamini and Yosef Hochberg.
\newblock Controlling the false discovery rate: A practical and powerful
  approach to multiple testing.
\newblock 2007.

\bibitem[Bubeck and Cesa-Bianchi(2012)]{Bubeck2012RegretAO}
S{\'e}bastien Bubeck and Nicol{\`o} Cesa-Bianchi.
\newblock Regret analysis of stochastic and nonstochastic multi-armed bandit
  problems.
\newblock \emph{Foundations and Trends in Machine Learning}, 5:\penalty0
  1--122, 2012.

\bibitem[Bubeck et~al.(2009)Bubeck, Munos, and Stoltz]{Bubeck2009PureEI}
S{\'e}bastien Bubeck, R{\'e}mi Munos, and Gilles Stoltz.
\newblock Pure exploration in multi-armed bandits problems.
\newblock In \emph{ALT}, 2009.

\bibitem[Carpentier and Valko(2015)]{Carpentier2015SimpleRF}
Alexandra Carpentier and Michal Valko.
\newblock Simple regret for infinitely many armed bandits.
\newblock In \emph{ICML}, 2015.

\bibitem[Chandrasekaran and Karp(2012)]{Chandrasekaran2012FindingTM}
Karthekeyan Chandrasekaran and Richard~M. Karp.
\newblock Finding the most biased coin with fewest flips.
\newblock \emph{CoRR}, abs/1202.3639, 2012.

\bibitem[Chaudhuri and Kalyanakrishnan(2017)]{chaudhuri2017pac}
Arghya~Roy Chaudhuri and Shivaram Kalyanakrishnan.
\newblock Pac identification of a bandit arm relative to a reward quantile.
\newblock In \emph{AAAI}, pages 1777--1783, 2017.

\bibitem[Chernoff(1959)]{chernoff1959sequential}
Herman Chernoff.
\newblock Sequential design of experiments.
\newblock \emph{The Annals of Mathematical Statistics}, 30\penalty0
  (3):\penalty0 755--770, 1959.

\bibitem[Deng et~al.(2013)Deng, Xu, Kohavi, and Walker]{deng2013improving}
Alex Deng, Ya~Xu, Ron Kohavi, and Toby Walker.
\newblock Improving the sensitivity of online controlled experiments by
  utilizing pre-experiment data.
\newblock In \emph{Proceedings of the sixth ACM international conference on Web
  search and data mining}, pages 123--132. ACM, 2013.

\bibitem[Deng et~al.(2016)Deng, Lu, and Chen]{Deng2016ContinuousMO}
Alex Deng, Jiannan Lu, and Shouyuan Chen.
\newblock Continuous monitoring of a/b tests without pain: Optional stopping in
  bayesian testing.
\newblock \emph{2016 IEEE International Conference on Data Science and Advanced
  Analytics (DSAA)}, pages 243--252, 2016.

\bibitem[Deng et~al.(2017)Deng, Lu, and Litz]{Deng2017TrustworthyAO}
Alex Deng, Jiannan Lu, and Jonthan Litz.
\newblock Trustworthy analysis of online a/b tests: Pitfalls, challenges and
  solutions.
\newblock In \emph{WSDM}, 2017.

\bibitem[Foster and Stine(2007)]{Foster2007AlphainvestingAP}
Dean~P. Foster and Robert~A. Stine.
\newblock Alpha-investing: A procedure for sequential control of expected false
  discoveries.
\newblock 2007.

\bibitem[Freeman(1983)]{freeman1983secretary}
PR~Freeman.
\newblock The secretary problem and its extensions: A review.
\newblock \emph{International Statistical Review/Revue Internationale de
  Statistique}, pages 189--206, 1983.

\bibitem[Gittins et~al.(2011)Gittins, Glazebrook, and Weber]{gittins2011multi}
John Gittins, Kevin Glazebrook, and Richard Weber.
\newblock \emph{Multi-armed bandit allocation indices}.
\newblock John Wiley \& Sons, 2011.

\bibitem[Jamieson et~al.(2014)Jamieson, Malloy, Nowak, and
  Bubeck]{Jamieson2014lilU}
Kevin~G. Jamieson, Matthew Malloy, Robert~D. Nowak, and S{\'e}bastien Bubeck.
\newblock lil' ucb : An optimal exploration algorithm for multi-armed bandits.
\newblock In \emph{COLT}, 2014.

\bibitem[Jamieson et~al.(2016)Jamieson, Haas, and Recht]{Jamieson2016ThePO}
Kevin~G. Jamieson, Daniel Haas, and Benjamin Recht.
\newblock The power of adaptivity in identifying statistical alternatives.
\newblock In \emph{NIPS}, 2016.

\bibitem[Javanmard and Montanari(2016)]{Javanmard2016OnlineRF}
Adel Javanmard and Andrea Montanari.
\newblock Online rules for control of false discovery rate and false discovery
  exceedance.
\newblock \emph{CoRR}, abs/1603.09000, 2016.

\bibitem[Johari et~al.(2017)Johari, Koomen, Pekelis, and
  Walsh]{Johari2017PeekingAA}
Ramesh Johari, Pete Koomen, Leonid Pekelis, and David Walsh.
\newblock Peeking at a/b tests: Why it matters, and what to do about it.
\newblock In \emph{KDD}, 2017.

\bibitem[Kaufman et~al.(2017)Kaufman, Pitchforth, and
  Vermeer]{kaufman2017democratizing}
Raphael~Lopez Kaufman, Jegar Pitchforth, and Lukas Vermeer.
\newblock Democratizing online controlled experiments at booking.com.
\newblock \emph{arXiv preprint arXiv:1710.08217}, 2017.

\bibitem[Kharitonov et~al.(2015)Kharitonov, Vorobev, MacDonald, Serdyukov, and
  Ounis]{Kharitonov2015SequentialTF}
Eugene Kharitonov, Aleksandr Vorobev, Craig MacDonald, Pavel Serdyukov, and
  Iadh Ounis.
\newblock Sequential testing for early stopping.
\newblock 2015.

\bibitem[Kohavi et~al.(2009)Kohavi, Crook, Longbotham, Frasca, Henne, Ferres,
  and Melamed]{Kohavi2009OnlineEA}
Ronny Kohavi, Thomas Crook, Roger Longbotham, Brian Frasca, Randy Henne,
  Juan~Lavista Ferres, and Tamir Melamed.
\newblock Online experimentation at microsoft.
\newblock 2009.

\bibitem[Lai et~al.(2011)Lai, Poor, Xin, and Georgiadis]{Lai2011QuickestSO}
Lifeng Lai, H.~Vincent Poor, Yan Xin, and Georgios Georgiadis.
\newblock Quickest search over multiple sequences.
\newblock \emph{IEEE Transactions on Information Theory}, 57:\penalty0
  5375--5386, 2011.

\bibitem[Lai(1988)]{lai1988nearly}
Tze~Leung Lai.
\newblock Nearly optimal sequential tests of composite hypotheses.
\newblock \emph{The Annals of Statistics}, pages 856--886, 1988.

\bibitem[Lai(1997)]{lai1997optimal}
Tze~Leung Lai.
\newblock On optimal stopping problems in sequential hypothesis testing.
\newblock \emph{Statistica Sinica}, 7\penalty0 (1):\penalty0 33--51, 1997.

\bibitem[Locatelli et~al.(2016)Locatelli, Gutzeit, and
  Carpentier]{Locatelli2016AnOA}
Andrea Locatelli, Maurilio Gutzeit, and Alexandra Carpentier.
\newblock An optimal algorithm for the thresholding bandit problem.
\newblock In \emph{ICML}, 2016.

\bibitem[Malloy et~al.(2012)Malloy, Tang, and Nowak]{Malloy2012QuickestSF}
Matthew Malloy, Gongguo Tang, and Robert~D. Nowak.
\newblock Quickest search for a rare distribution.
\newblock \emph{2012 46th Annual Conference on Information Sciences and Systems
  (CISS)}, pages 1--6, 2012.

\bibitem[Page(1954)]{page1954continuous}
Ewan~S Page.
\newblock Continuous inspection schemes.
\newblock \emph{Biometrika}, 41\penalty0 (1/2):\penalty0 100--115, 1954.

\bibitem[Ramdas et~al.(2017)Ramdas, Yang, Wainwright, and
  Jordan]{ramdas2017online}
Aaditya Ramdas, Fanny Yang, Martin~J Wainwright, and Michael~I Jordan.
\newblock Online control of the false discovery rate with decaying memory.
\newblock In \emph{Advances in Neural Information Processing Systems}, pages
  5655--5664, 2017.

\bibitem[Russo(2016)]{russo2016simple}
Daniel Russo.
\newblock Simple bayesian algorithms for best arm identification.
\newblock In \emph{Conference on Learning Theory}, pages 1417--1418, 2016.

\bibitem[Samuels(1991)]{samuels1991secretary}
Stephen~M Samuels.
\newblock Secretary problems.
\newblock \emph{Handbook of sequential analysis}, 118:\penalty0 381--405, 1991.

\bibitem[Shiryayev(1978)]{shiryayev1978optimal}
Alexi~N Shiryayev.
\newblock Optimal stopping rules, volume 8 of applications of mathematics,
  1978.

\bibitem[Siegmund(2013)]{siegmund2013sequential}
David Siegmund.
\newblock \emph{Sequential analysis: tests and confidence intervals}.
\newblock Springer Science \& Business Media, 2013.

\bibitem[Tang et~al.(2010{\natexlab{a}})Tang, Agarwal, O'Brien, and
  Meyer]{google_experiments}
Diane Tang, Ashish Agarwal, Deirdre O'Brien, and Mike Meyer.
\newblock Overlapping experiment infrastructure: More, better, faster
  experimentation.
\newblock In \emph{Proceedings 16th Conference on Knowledge Discovery and Data
  Mining}, pages 17--26, Washington, DC, 2010{\natexlab{a}}.

\bibitem[Tang et~al.(2010{\natexlab{b}})Tang, Agarwal, O'Brien, and
  Meyer]{google_experiments_preso}
Diane Tang, Ashish Agarwal, Deirdre O'Brien, and Mike Meyer.
\newblock Overlapping experiment infrastructure: More, better, faster
  experimentation (presentation).
\newblock 2010{\natexlab{b}}.
\newblock URL
  \url{https://static.googleusercontent.com/media/research.google.com/en//archive/papers/Overlapping_Experiment_Infrastructure_More_Be.pdf}.

\bibitem[Wald and Wolfowitz(1948)]{wald1948optimum}
Abraham Wald and Jacob Wolfowitz.
\newblock Optimum character of the sequential probability ratio test.
\newblock \emph{The Annals of Mathematical Statistics}, pages 326--339, 1948.

\bibitem[Wetherill and Glazebrook(1986)]{wetherill1986sequential}
G~Barrie Wetherill and Kevin~D Glazebrook.
\newblock Sequential methods in statistics, 1986.

\bibitem[Williams(1991)]{williams1991probability}
David Williams.
\newblock \emph{Probability with martingales}.
\newblock Cambridge university press, 1991.

\bibitem[Yang et~al.(2017)Yang, Ramdas, Jamieson, and
  Wainwright]{yang2017framework}
Fanny Yang, Aaditya Ramdas, Kevin~G Jamieson, and Martin~J Wainwright.
\newblock A framework for multi-a (rmed)/b (andit) testing with online fdr
  control.
\newblock In \emph{Advances in Neural Information Processing Systems}, pages
  5959--5968, 2017.

\end{thebibliography}

    \newpage
    \appendix

\section{Proofs}

\subsection{Proofs from Section~\ref{sec:model}}

\begin{proof}[Proof of Lemma~\ref{thm:validity}]
  The result relies on $\tau$ being a stopping time.
  Recall that $i^*$ indicates the discovered experiment.
  Then we find
  \begin{align*}
        \P(\mu_{i^*} < s \mid \cF_\tau)
        &= \sum_{t=1}^\infty \P(\mu_i^* < s \mid \cF_\tau \cap \{\tau = t\} ) \P(\tau=t)\\
        &= \sum_{t=1}^\infty \P(\mu_i^* < s \mid \cF_t ) \P(\tau=t)\\
        &\le \alpha \sum_{t=1}^\infty \P(\tau=t) = \alpha
  \end{align*}
  where we use that
  $F \in \cF_\tau$ if $F \cap \{\tau = t\} \in \cF_t$, and thus
  $\cF_\tau = \cF_t$ if $\tau = t$ \citep{williams1991probability}[p.219].
\end{proof}

\begin{proof}[Proof of Lemma~\ref{thm:sequential}]
    Note that due to independence we can assume without loss of generality that
    the index of the arm corresponds to the order in which alternatives are
    first considered.  Thus the result follows if we show that for any $t$,
    action $I_t < I_{t-1}$ cannot be strictly better than $I_t = I_{t+1}$.
    Assume to the contrary that $I_t = y$ is optimal (and strictly better than
    $I_t = I_{t-1} + 1$ for some $y < I_t$.  Consider the last time alternative $y$
    was selected: $t' = \max \{k < t:  I_k = y\}$. At that time it was at least
    as good to consider a new alternative, and subsequently the posterior
    for alternative $y$ has not changed due to independence.
    Due to the infinite time horizon, it is thus at least as good to consider
    a new alternative.
\end{proof}

\subsection{Proofs from Section~\ref{sec:optimal}}

\begin{proof}[Proof of Lemma~\ref{thm:acceptance}]
  Let $n \ge 1$.
  We can rewrite the discovery criterion as
  \begin{align}
    \P(\mu < s \mid S_n = t)
    &= \frac{ \int_{-\infty}^s \prod_{i=1}^n h(X_i) \exp \left( \mu S(X_i) - A(\mu) \right) d\pi(\mu)}
            { \int_{-\infty}^\infty \prod_{i=1}^n h(X_i) \exp \left(  \mu S(X_i) - A(\mu) \right) d\pi(\mu)}\\
    &= \frac{ \int_{-\infty}^s \exp \left( \mu S_n - nA(\mu) \right) d\pi(\mu)}{ \int_{-\infty}^\infty \exp \left( \mu S_n - nA(\mu) \right) d\pi(\mu)}\\
    &= \frac{ \int_{-\infty}^s \exp \left( \mu t - nA(\mu) \right) d\pi(\mu)}{ \int_{-\infty}^\infty \exp \left( \mu t - nA(\mu) \right) d\pi(\mu)}
  \end{align}
  We show that this is decreasing in $t$.

  Now take the logarithm and the derivative with respect to $t$ to obtain
  Find expression
  \begin{align}
    \frac{d}{dt} \log(\P(\mu < s \mid S_n = t))
    &= \frac{\int_{-\infty}^s \mu \exp(\mu t - nA(\mu)) d\pi(\mu)}{\int_{-\infty}^s \exp(\mu t - nA(\mu)) d\pi(\mu)}\\
    &\quad- \frac{\int_{-\infty}^\infty \mu \exp(\mu t - nA(\mu)) d\pi(\mu)}{\int_{-\infty}^\infty \exp(\mu t - nA(\mu)) d\pi(\mu)}\\
    &= \E_{f_t}(\mu \mid \mu < s) - \E_{f_t}(\mu) < 0
  \end{align}
  where the expectations in the last line is taken with respect to the distribution with density
  \[
    f_t(\mu) = \frac{\exp(\mu t - nA(\mu)) d\pi(\mu)}{\int_{-\infty}^s \exp(\mu t - nA(\mu)) d\pi(\mu)}\\
  \]
  Note that the last inequality holds, because, in general
  \begin{multline}
    \E(\theta) = \E(\theta \mid \theta < s) \P(\theta < s) + \E(\theta \mid \theta \ge s) \P(\theta \ge s)\\
    > \E(\theta \mid \theta < s) \P(\theta < s) + s \P(\theta \ge s) > \E(\theta | \theta < s)
  \end{multline}
  Now the lemma follows: if $\P(\mu <s \mid S_n = t) < \alpha$, then $\P(\mu <s
  \mid S_n = t') < \alpha$ for all $t' > t$, and similarly if if $\P(\mu <s
  \mid S_n = t) > \alpha$, then $\P(\mu <s \mid S_n = t') > \alpha$ for all $t'
  < t$.
\end{proof}

To prove the theorems in Section~\ref{sec:optimal} we use the following lemmas,
which are proven at the end of this section.

\begin{lemma}
    \label{thm:truncated_threshold}
    The optimal policy for the truncated problem can be characterized
    by a rejection threshold.
    That is, the optimal policy rejects the current experiment if $S_n < r^k_n$
    for a sequence $r^k_n$, and collects another observation for the current experiment
    otherwise, until a discovery is made.
\end{lemma}

Write $T_k$ for the expected number of observations required for a discovery
for the optimal policy of the truncated problem.
Then we can show that both $T_k$ and $r_n^k$ converge.

\begin{lemma}
    \label{thm:truncated_convergence}
    Both $T_k$ and $r_n^k$ converge as $k \to \infty$.
\end{lemma}

\begin{proof}[Proof of Theorem~\ref{thm:optimal}]
  Lemma~\ref{thm:truncated_threshold} shows that the truncated problem
  has an optimal policy that has the form of a threshold.
  Next, lemma~\ref{thm:truncated_convergence} shows that both the thresholds
  and the optimal cost converge.

  Recall $T^* = \lim_{k\to\infty} T_k^*$ and $r_n = \lim_{k\to\infty} r_n^k$.
  Now we show that limiting policy $r_n$ with corresponding cost $T^*$ is optimal.

  Suppose there exists an $\epsilon > 0$ and a policy $\b \phi$ with cost $\b T$ such
  that $\b T = T^* - \epsilon$.
  Consider a policy with cost $\b T < T^*$.
  Let $\b \tau$ be the stopping time of this policy.
  We consider the truncated version of this policy, and show that it
  cannot be much worse. On the other hand, this truncated policy
  has a cost larger than $T^*$.
  The $k$-truncated policy, denoted by $\b \phi_k$ rejects the current alternative after $k$ samples,
  but is otherwise identical to $\b \phi$.
  Let $\b \tau$ and $\b \tau_k$ be the stopping times corresponding to $\b \phi$ and $\b \phi_k$.
  Trivially, we have $\b T = \sum_{k=1}^\infty \P(\b \tau \ge k)$. Because $\b T$ is finite,
  $\P(\b \tau \ge k) = O((k \log k)^{-1})$.
  Because $\b \phi$ and $\b \phi_k$ are identical up to $k$ observations,
  it follows that if $\b \tau < k$, then $\b \tau_k < k$, and thus we find that
  \begin{align*}
      \E(\b \tau_k)
      &= \P(\b \tau > k) \E(\b \tau_k \mid \b \tau > k) +
          \E(\b \tau \ind(\b \tau \le k)) \\
      &\le \P(\b \tau > k) (k + \E(\b \tau_k))  +
          \E(\b \tau \ind(\b \tau \le k))
  \end{align*}
  Thus, it follows that
  \[
    \E(\b \tau_k) \le
      \frac{k \P(\b \tau > k)}{1-\P(\b \tau > k)}
      + \frac{\b T}{1 - \P(\tau > k)}.
  \]
  Since $\P(\b \tau > k) = O((k \log k)^{-1})$,
  $\E(\b \tau_k) \to \b T$ as $k \to \infty$.

  However, $ T^* \le \E(\b \tau_k) $ for all $k$, and thus $T^* \le \lim_{k\to\infty} \E(\b \tau_k) = \b T$,
  which is a contradiction.
\end{proof}

\begin{proof}[Proof of Theorem~\ref{thm:truncated}]
  This is a direct consequence of lemmas \ref{thm:truncated_threshold} and \ref{thm:truncated_convergence}.
\end{proof}

\begin{proof}[Proof of Theorem~\ref{thm:onestep}]
  Let $\tau_r$ denote the (random) hitting time of the boundary of the first alternative
  \[
      \tau_r = \min \{ n : S_n \ge a_n \text{ or } S_n < r_n \}
  \]
  under rejection boundary $r = \{r_n\}_{i=1}^k$.
  Furthermore, let $q_r = \P(S_\tau < r_\tau)$ denote the rejection
  probability.
  Now note that $f(\kappa) = \min_r \E(\tau_r) + \kappa q_r$
  Note that we can solve this minimization problem using backward induction, since the
  time horizon is fixed ($k$).
  First, we show that $f$ has a unique fixed point which is equal to $T_k^*$.

  Note that we have
  \[
      T_r = \E(\tau_r) + T_r q_r
      \label{eqn:Tr}
  \]
  By definition, $T_k^*$ minimizes $\min_r \E(\tau_r) + T_k^* q_r$, thus,
  it follows immediately that $T_k^*$ is a fixed point of $f$.

  Next, we show that $f(\kappa) > \kappa$ for each $\kappa < T_k^*$ and
  $f(\kappa) < \kappa$ for each $\kappa > T_k^*$.

  First, fix $\kappa < T_k^*$. Suppose that $f(\kappa) \le \kappa$. Thus, there exists $r'$ such
  that $\E(\tau_{r'}) + \kappa q_{r'} \le \kappa$.
  Thus, $\kappa \ge \frac{\E(\tau_{r'})}{1 - q_{r'}} = T_{r'}$, where the last
  equality follows from \eqref{eqn:Tr}. This, along with $\kappa < T_k^*$ implies that
  $T_{r'} < T_k^*$, a contradiction. Thus, we must have $f(\kappa) > \kappa$.

  Finally, fix $\kappa > T^*$.
  We know that
  \[
      T^* = \E(\tau_{r^*}) + T_k^* q_{r^*} < \E(\tau_{r^*}) + \kappa q_{r^*}.
  \]
  Thus, there exists
  $r$ (equal to $r^*$) such that $\E[\tau_{r}] + \kappa q_r < \kappa$.
  Thus, $f(\kappa) < \kappa$.
\end{proof}

\subsection{Proofs of lemmas}

\begin{proof}[Proof of Lemma~\ref{thm:truncated_threshold}]

 Based on Lemma~\ref{thm:sequential}, there exists a policy that can be characterized
 by a sequence of three sets
 \begin{itemize}
   \item \emph{Discover} if $S_n \in A_n$, the experiment is a discovery
   \item \emph{Continue} if $S_n \in D_n$, and
   \item \emph{Reject} if $S_n \in R_n$
 \end{itemize}

Now note that $R^k_n$ is a threshold region for $n \ge k$ by definition.
 Assume $R^k_{m} = (-\infty, r_{m}^k]$ for all $m > n$.
Further, from the Bellman equation for the truncated problem, it is clear that the optimal solution rejects the  current experiment at the time $n$ if
\[
  \E[V_k(n+1,S_{n+1})\mid S_n]  > \E[V_k(1,S_1)]  = T_{k}^* - 1.
\]

Note that
 \begin{multline}
\E[V_k(n+1,S_{n+1}) \mid S_n = x] =
  \int_{y \in D_{n+1}} V_k(n+1,y) f(S_{n+1} = y \mid S_n = x) dy\\
  + T_k^* \P(S_{n+1} < r_{n+1} \mid S_n = x)
\end{multline}

Then for $n$ we note that $\E[V_k(n+1,S_{n+1}) \mid S_n = x] $ is decreasing in $x$.
This follows since $V_k(n+1, y) < T_k^*$ for all $y \in D_{n+1} = [r_{n+1}, a_{n+1}]$,
as for such $y$ it is better to continue than to reject. Furthermore,
arguing along the lines of the proof of Lemma~\ref{thm:acceptance}, $\P(S_{n+1} <
r_{n+1} \mid S_n = x)$ is decreasing in $x$.
 This implies we can write $R_n^k = (-\infty, r_n^k]$ for some $r_n^k$.
 \end{proof}

\begin{proof}[Proof of Lemma~\ref{thm:truncated_convergence}]
    Due to increased degrees of freedom, it follows that $T_k^*$ is decreasing.
    Since $T_k^*$ is bounded below by $0$, $T_k^n$ converges.
    Let $T^* = \lim_k T_k^*$.

    Next, we show that $r_n^k$ is decreasing in $k$:
    Clearly, $r_k^{k+1} \le r_k^k$. Now suppose
    $r_{n+1}^{k+1} \le r_{n+1}^k$, then $r_n^{k+1} \le r_n^k$,
    which follows from the fact that $\E[V_k(n+1, S_{n+1}) \mid S_n = r_n^k] + 1= T_k^*$, and
    $T_k^*$ is decreasing in $k$.
    It remains to show that $r_n^k$ is bounded.

    We construct a lower bound on $r_n^k$, for large $k$, as follows.
    Let $\epsilon = \frac{1}{2T^*}$ and let
    $x$ be such that $\P(\exists m > n \text{ s.t. } S_m > a_m \mid S_n = x) < \epsilon$, by
    choosing $x$ sufficiently small.
    Then we note that the cost for obtaining another sample is at least
    $1 + (1-\epsilon) T_k^* \ge 1 + (1-\epsilon) T^* = T^* + 1/2$.
    However, if the experimenter rejects the current alternative now,
    the cost is $T_k^*$.
    Thus, if we can show that there exists a $K$ such that for all $k > K$,
    $T_k^* < T^* + 1/2$, then $x$ is a lower bound on $r_n^k$ for all $k > K$.
    But above we have shown that $T_k^* \to T^*$, hence such $K$ exists.
    This implies that $r_n^k$ converges as $k\to\infty$.
\end{proof}

\end{document}